\date{January 2024}
\documentclass{article}
\usepackage{spconf,amsmath,graphicx, subcaption}
\usepackage{hyperref}
\usepackage{amsfonts}

\title{Fill in the blanks: Rethinking Interpretability in vision}
\name{Pathirage N. Deelaka*, Tharindu Wickremasinghe*, Devin Y. De Silva*, Lisara N. Gajaweera \textsuperscript{$\dagger$}}
\address{ *University of Moratuwa, Sri Lanka \\ 
\textsuperscript{$\dagger$} Robert Gordon University, Scotland}
\begin{document}
%
\maketitle
\begin{abstract}
Model interpretability is a key challenge that has yet to align with the advancements observed in contemporary state-of-the-art deep learning models. In particular, deep learning aided vision tasks require interpretability, in order for their adoption in more specialized domains such as medical imaging. Although the field of explainable AI (XAI) developed methods for interpreting vision models along with early convolutional neural networks, recent XAI research has mainly focused on assigning attributes via saliency maps. As such, these methods are restricted to providing explanations at a sample level, and many explainability methods suffer from low adaptability across a wide range of vision models. In our work, we re-think vision-model explainability from a novel perspective, to probe the general input structure that a model has learnt during its training. To this end, we ask the question:``How would a vision model fill-in a masked-image". Experiments on standard vision datasets and pre-trained models reveal consistent patterns, and could be intergrated as an additional model-agnostic explainability tool in modern machine-learning platforms. The code will be available at \url{https://github.com/BoTZ-TND/FillingTheBlanks.git}

\end{abstract}

\section{Introduction}
\label{sec:intro}

The rapid advancements of Machine Learning (ML) and Deep Learning(DL) models show no signs of slowing down, with multimodal models further alluring researchers into pushing performance limits. However, the challenge of adopting DL architectures into critical applications such as medical imaging is largely due to our limited understanding of the inner workings of models. In an effort to understand these ``black box" models, the field of explainable AI (XAI) emerged with a special focus on \textit{`interpretability'}, and \textit{`explainability'}.
Although these terms are sometimes used interchangeably, following efforts to define these terms more concretely \cite{MontavonMethods}, we shall define them for our work in the context of vision models. \textit{`Interpretability'} is when a concept such as a model prediction, is mapped back to an image that makes sense to a human. It answers the question ``What input would generate a given output/ activation?". \textit{`Explainability'} aims to find the contribution of image features towards the model's output. In other words ``What features of this image were significant for the model's decision?".

The former was the main focus among researchers when convolutional neural network (CNN) architectures were being developed for vision tasks. Particularly focused on classification, models \cite{Deconvnet_Zeiler2013,UnderstandingInverting_Mahendran2014} were trained to predict to search for input images that maximized a certain activation node in the network. Consequently, input features which contributed to a particular class were visualised in the image space. We shall refer to this as a `global' understanding of what the model has learnt as the concept of a given class. However, as we shall elaborate in section \ref{sec:relatedwork}, these methods involved a separate model to be trained, and hence is impractical for fast inference on new models. 

A more `local' approach developed, where given an input image and an activation, each pixel was scored corresponding to its contribution towards that activation \cite{understandvisualising_yosinski2015}. These patterns were visualized using heat-maps to reveal the pixel patterns within the input that an activation/node is particularly sensitive to. Therefore, the focus was on \textit{explaining} the input, rather than \textit{interpreting} what the model had learnt as a whole. In the context of an image classifier, this meant observing which pixels made the model choose a given class (e.g: digit `three'), but not observing what the model generally perceives as `three'. This would not give an idea on what is the `prototypical digit' that the model learnt after training on samples from an image distribution.

In this light, we strive to re-visit the global approach of visually interpreting vision models. However, instead of generating a prototypical image, we ask a different question: ``How would a vision model fill-in a masked image?". Ideally, we expect to visualise how the model would fill in masked out portions of an image, based on the limited priors that the unmasked patches provide.
We answer this in the context of image classification, and propose a method that does not require to train a separate image generator.  Through this novel approach of interpreting image classifiers, we present the following:
\begin{enumerate}
    \item Introduce a mask-filling approach for visually interpreting vision models, as an alternative to existing generative methods.
    \item Present different approaches of masking, and interpret the visual results on standard data sets.
    \item Demonstrate consistency of the visual patterns and present the effects of changing masking parameters.
\end{enumerate}

\section{Related Work}
\label{sec:relatedwork}

Our work aims to interpret learned features of an image classifier based on a visual output. The conclusions on the model will be interpreted by the human, based on perceived patterns of the visual result. Therefore, we present a brief overview of literature that can be considered within XAI methods of `perceptive interpretability'. 

\subsection{Saliency maps}
These methods aim to assign an attribute (common choices include relevance/saliency/importance) for each pixel in the input image, which corresponds to the impact the pixel has on the prediction of the model. More formally, suppose our prediction from model $f$ is $f(x)$. Then the relavance $R(x_{i})$ will be a scalar value assigned for each pixel $x_{i}$. 
Different choices of attributing pixels range from introducing `relevance score' \cite{LRP_Bach2015}, to separating out the superpixels that classify the input image as it's predicted class \cite{LIME_Ribeiro2016}. In \cite{DEEPLIFT_Shrikumar2017}, $R(x_{i})$ represents the effect of that input pixel being set to a reference value as opposed to its original value. Developing on these ideas, \cite{AxiomaticAF_Sundararajan2017} introduced `sensitivity' and `implementation invariance' as axioms that an attribution score $R$ should satisfy. Another approach has been Class Activation Maps (CAM)\cite{CAM_Zhou2015}, in which Zhou et.al visualized discriminative regions for action classification on a CNN trained on another task. This lead to a family of models including Grad-CAM\cite{GradCAM_Selvaraju2016}, Score-CAM\cite{ScoreCAM_Wang2019}, Ablation-CAM\cite{AblationCAM_Desai2020}, and Eigen-CAM\cite{EigenCAM_Fu2020}. 

These `local' methods are adaptable to many vision models that conform to some structure. For example, GradCAM \cite{GradCAM_Selvaraju2016} is claimed to be applicable to \textit{any} CNN-based vision model. We also note that the metrics that are visualized as saliency maps are derived from gradients that flow from a particular node/activation of the network. Thereby, they are easily implemented from existing frameworks such as pytorch and tensorflow.  However, since these methods aim at highlighting regions of a specific input image that are more relevant for it's classification, they do not provide a global view of the discriminative features that the model has learnt about that class. 

To answer question ``How would a vision model fill in masked-out patches?", we build upon the strengths of the above methods. Eventhough the aim is not to find a pixel-wise attribution score, our method leverages the unmasked regions of an input as an instance-level prior. We generate a visualisation for the masked out regions (alluding to the `global' features learnt by the model), while each fill is conditioned on the unmasked features that are `local' for each input image. Additionally, our visual output is reproducible for many networks and only requires a gradient calculation with respect to a classification loss. It does not necessitate a specific layer (such as a CNN) to exist before the final logits of a classifier.

\subsection{Generating a representative input}
Early work on interpreting features of deep networks employed gradients to visualise prototypical features corresponding to a particular activation/node. For example, \cite{Erhan2009VisualizingHF} used gradient descent in image space to maximize a unit’s activation. Others (e.g: Deconv net\cite{Deconvnet_Zeiler2013}) strived to invert basic structures of CNN models  to interprit activations of a particular class. Simonyan et. al \cite{DeepInside_Simonyan2013} generalized the Deconv net as a special case of visualising the gradient through backpropagation. Guided backpropagation \cite{StrivingFS_Springenberg2014} was another attempt at modifying backpropagation of gradients to accommodate better visualizations in the image space. 
However, visualising invariant features of a particular image class was the popular method for interpreting a model, and gradient-based methods were empirically shown to be poor in this regard. Subsequently, research focused on solving for a `prototypical image' though an optimization in the image space \cite{UnderstandingInverting_Mahendran2014}.

As noted by \cite{VisualCNNPreImg_Mahendran2015}, searching the input space to visualise invariant features led to formalising the search as an `activation maximization', an `inversion', or a `caricaturisation' of a particular class. With the contemporary traction gained by generative models, the task of searching for a prototypical image evolved into learning rich embeddings from latent spaces (e.g: \cite{SynthesizingTP_Nguyen2016}, \cite{PPGN_Nguyen2017} )  
rather than optimizing directly in the image space. A main drawback of these methods is the requirement to train a separate inverter/generator. Furthermore, since the inverters are themselves DL models, it is not straightforward to interpret a generated image as features learnt by the original vision model.

Interestingly, we see that for the task of filling in masked patches, employing a gradient-based method provides interpretable visual outputs. Given the priors from unmasked patches, generating prototypical patterns for the masked regions need not employ a separately trained DL model.

\section{Methodology}
\label{sec:method}

\begin{figure}[htb]
\begin{minipage}[b]{1.0\linewidth}
  \centering
  \centerline{\includegraphics[width=8.5cm]{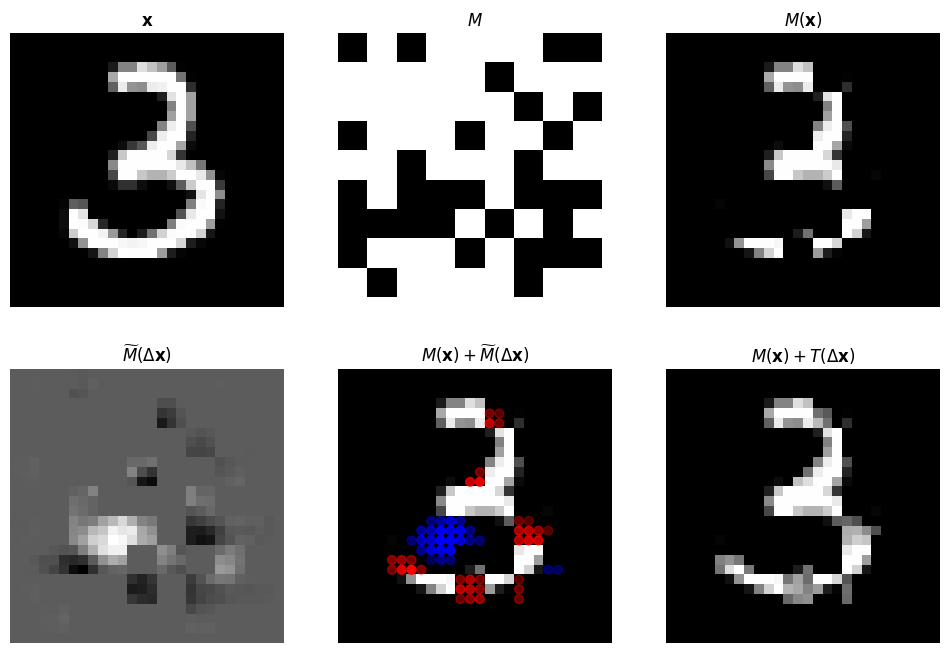}}
\end{minipage}
\caption{Evaluating a simple CNN classifier trained on MNIST\cite{lecun2010mnist}. Mask $M$ is applied on image $\mathbf{x}$ of the digit 3. The positive (red) and negative (blue) components of the update term $\widetilde{M}(\mathbf{x})$ is shown superimposed with $M(\mathbf{x})$. The thresholded update $T( \Delta \mathbf{x} )$ `fills in the blanks' of the masked input. }
\label{fig:method}
\end{figure}

\subsection{Rethinking interpretability}

Let us focus on the task of image classification, and re-visit how a human recalls from memory. When an image, say of the number `three', is shown to a human, we associate that image with previous memories of the digit, and identify distinctive features of the digit. However, this mental image of a digit is not unique for a human mind. There may be many such digits that a person recalls, that have variances in size, tilt, font...etc. Consequently, as observed by \cite{olah2017feature}, searching for an image that maximized a single activation generates an image that a human would not perceive as a `three'. Therefore we search for alternatives for interpreting distinctive features learnt by a classifier.

With this motivation, we pose the following task: Let the model predict the class of a masked-out image of the digit `three'. Depending on what the model has learnt about the digit `three' in general, it will generate a prediction.
If it generates the prediction correctly, we ask the model to generate a suggestion for the masked-out patches.
Visualising this in comparison to the original image, we can perceive what the model has learnt as distinctive features that lead to it's classification. Ideally, we aim to visualise a digit, which was pre-conditioned on the unmasked regions. This allows the model to be interpreted by repeating this process on many different images, and it would not generate obscure images as seen from maximizing activations.

In the following, we present our approach of probing an image classifier $f$ in two main steps: 1) Predicting the class for a masked image, and 2) Visualising how $f$ tends to `fill-in' the masked patches, as a way of interpreting ``How $f$ generates the prototypical class based on the given structure".

\subsection{Formalisms}

Let an input image be the vector $\mathbf{x} \in \mathbb{R}^n$, where $n$ is the number of pixels in the image. Given $k$ classes, the parameters $\mathbf{\theta}$ of the image classifier $f_{\mathbf{\theta}}$ are trained to predict a probability vector $\hat{\mathbf{y}} = f_{\mathbf{\theta}}(\mathbf{x}) \in \mathbb{R}^k$, which is evaluated by some loss function $\mathcal{L} (\hat{\mathbf{y}},\mathbf{y})$. During training, $\mathbf{\theta}$ is updated using some form of stochastic gradient descent, which uses the update vector $ \Delta \mathbf{\theta} = \nabla_{\mathbf{\theta}} \mathcal{L} (f(\mathbf{x}),\mathbf{y})$.

However, to interpret the model $f$, we revisit the idea of treating the input $\mathbf{x}$  as a parameter to be optimized \cite{Erhan2009VisualizingHF}. Here, the idea is to find an update vector $\Delta \mathbf{x}$ such that at $\mathbf{x}_{t+1} = \mathbf{x}_{t} + \Delta \mathbf{x}$. Assuming smoothness of $\mathcal{L} (f(\mathbf{x}),\mathbf{y})$ at $\mathbf{x}= \mathbf{x}_{t}$, the update corresponds to a lower loss, and a closer prediction to the ground truth $\mathbf{y}$. Therefore, $\Delta \mathbf{x}$ is a ``suggestion" on how to change $\mathbf{x}_t$ for better prediction. Now suppose $\Delta \mathbf{x}$ is generated from $f_{\mathbf{\theta}}$ itself. From the perspective of $f_{\mathbf{\theta}}$, this is ``which update vector would make $\mathbf{x}_{t+1}$ more indicative of an image labelled as $\mathbf{y}$ ". Hence, visualising $\Delta \mathbf{x}$ is a method to interpret what the model has learnt as image features that would improve it's current prediction.

\textbf{Filling-in masked patches} is proposed as a method for interpreting what the model has learnt as prototypical features. Therefore, let us define $M(\mathbf{x})$ as our masked image for some binary masking function $M$. This corrupted image probes what the model predicts as it's perceived class as 
\begin{equation}
    \hat{\mathbf{y}} = f_{\mathbf{\theta}}(M(\mathbf{x})).
\end{equation}
Following the prediction, we calculate an update vector $\Delta \mathbf{x}$ by viewing the space $\mathbb{R}^n$ as a parameter space. Assuming $\mathcal{L} (f(\mathbf{x}),\mathbf{y})$ is locally smooth at $\mathbf{x}= \mathbf{x}_{t}$, we employ gradient descent: 
\begin{equation}
\begin{aligned}
    \Delta \mathbf{x} &= - \gamma \nabla_{\mathbf{x}} \mathcal{L}_{\mathbf{\theta}}(M(\mathbf{x}), \mathbf{y}). \\
    \mathbf{x}_{t+1} &= \mathbf{x}_{t} + \Delta \mathbf{x}.
\end{aligned}
\end{equation} 

Note that a gradient descent update with a sufficiently small step size $\gamma$ satisfies $\mathcal{L}_{\mathbf{\theta}}(M(\mathbf{x_{t+1}}), \mathbf{y}) < \mathcal{L}_{\mathbf{\theta}}(M(\mathbf{x_{t}}), \mathbf{y})$. Furthermore, the choice of the first order gradient is solely for ease of computation. An update incorporating a higher order numerical optimization is possible, and we provide evidence for this in section \ref{sec:experiments}. More broadly, we note that the direction of $\Delta \mathbf{x}$ conveys the information on how the classifier $f$ `selects' a class for $M(\mathbf{x})$ , and how $f$ `suggests' to improve $M(\mathbf{x})$ as an image from that selected class.

Our focus on how the model $f$ tends to fill-in the blanks of the masked patches. As such, we apply a threshold $T(\Delta \mathbf{x})$ such that we retain information only at the positions corresponding to the patches that lost information from $M()$. We superimpose the masked input image with this update vector. As shown below, this `fills-in' the masked patches of $M(\mathbf{x})$. 
\begin{equation}
    M(\mathbf{x}) + T( \Delta \mathbf{x} ).
\end{equation}

\textbf{Thresholding} is applied at two stages. The first, depends on each position of the pixel.
We apply the complement of mask $M$ to the update vector $\Delta \mathbf{x}$, as $\widetilde{M}(\Delta \mathbf{x})$. This retains the incremental updates only at the masked regions of $M(\mathbf{x})$. The second part of the threshold attempts to improve the contrast of $\widetilde{M}(\Delta \mathbf{x})$, as an aid to visualise the patterns on the same range of values of the input image. There is an additional motivation to filter some noise that remains from the gradient computation from a deep neural network. Hence a normalisation followed by a binary thresholding, is applied on $M(\mathbf{x})$ as follows.

\begin{equation}
    T( \Delta \mathbf{x} ) = Binarize( normalize ( \widetilde{M}(\Delta \mathbf{x}) ) ).
\end{equation}

Specific choices for the threshold transform are explored in section \ref{sec:experiments}. However, this choice would depend on what one would prefer to highlight in the visualisation. For instance, binary thresholding highlights patterns from the most predominant class. To visualise more general patterns, one could omit such thresholds.

In essence, our method visualizes the update vectors generated by the model, given the context of the masked image. By filling the blanks and comparing to the original image, we visually evaluate how well the model has learnt prototypical features of the class. If the patches that are filled aligns with the unmasked structure as shown in figure \ref{fig:method}, $f$ has learnt to `fill-in the blanks'.

\section{Experiments}
\label{sec:experiments}

\begin{figure}[htbp]
\centering
\begin{subfigure}[b]{0.45\textwidth}
  \includegraphics[width=\textwidth]{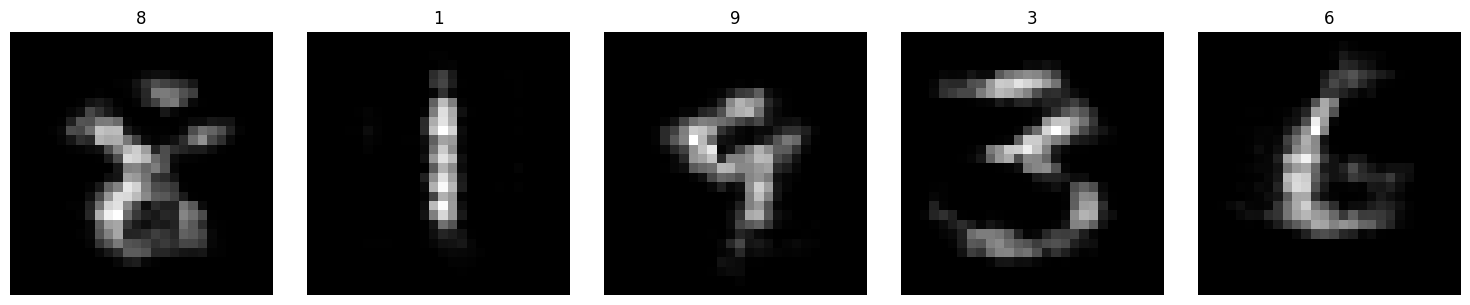}
  \caption{ }
\end{subfigure}
\hfill
\begin{subfigure}[b]{0.45\textwidth}
  \includegraphics[width=\textwidth]{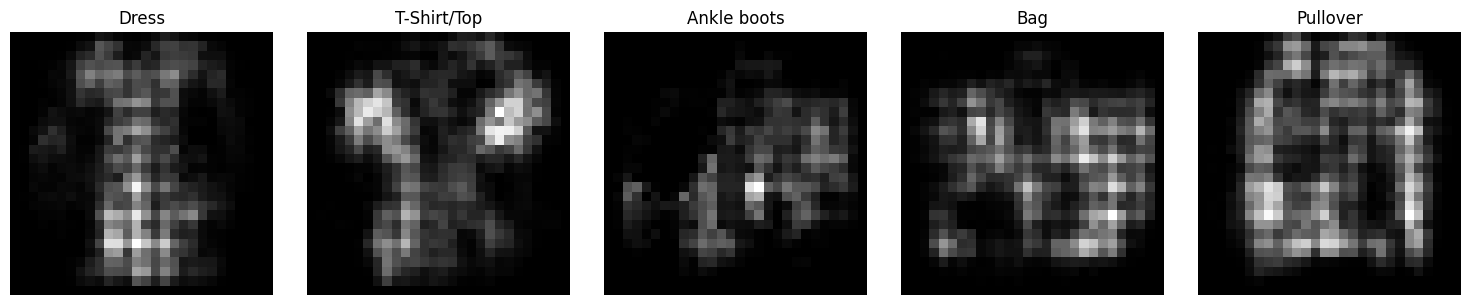}
  \caption{ }
\end{subfigure}

\caption{Visualising approximations of $\mathbb{E} [ \Delta \mathbf{x} ]$ for digits of (a) MNIST\cite{lecun2010mnist} and (b) Fashion MNIST\cite{Xiao2017FashionMNISTAN} data (section 4.1). Observe that the prototypical image of each class is visually convincing of distinctive features of that class.}
\label{fig:generalmask}
\end{figure}

\begin{figure}[htb]
\begin{minipage}[b]{1.0\linewidth}
  \centering
  \centerline{\includegraphics[width=8.5cm]{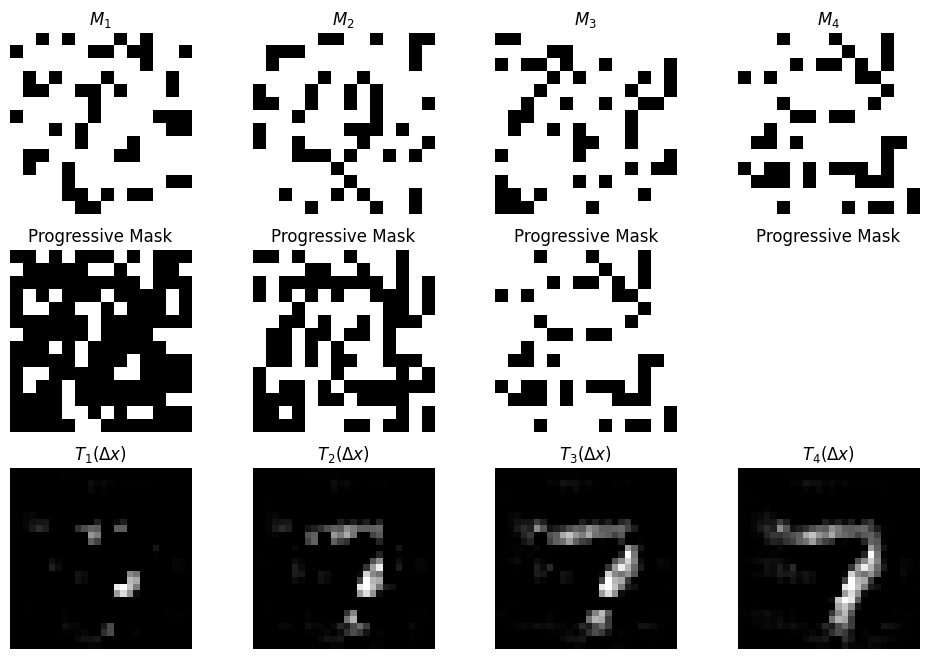}}
\end{minipage}
\caption{Visualising the predictions of progressively masking different regions of the image (section 4.2). The sequence $\mathcal{M}$ in this experiment has 4 non-overlapping masks, each masking 25\% of the pixels. Progressively the prediction of each mask is used to generate an update for all pixels.}
\label{fig:progressivemask}
\end{figure}

In this section, we present two ways of structuring the masking patterns to probe the prototypical image features associated with a class. Next, we experiment the consistency of such features across datasets, and models. Finally, we ablate our choices in our visualisations, such as the mask percentage, the patch size of the block, a second order update vector, and the threshold parameters of the update vector.

\subsection{$M$ as a uniform random variable}
Given an image of $n$ pixels, and a masking ratio $\eta$, suppose we randomly sample $M$ from a distribution of all possible binary masks of ratio $\eta$. Thus, each sample $M$ has $\eta n$ elements that are zero, and the masking transform $M(\mathbf{x})$ is an element-wise product of the mask, and the image. Given that $M$ is a random variable, it can be deduced that $\Delta \mathbf{x}$ is a random variable. If $f$ has learnt some prototypical structure from a particular class the expected value of the update should resemble the underlying structure of the class assigned for an image. Therefore, given $\mathbf{x}$, we visualise $ T( \mathbb{E} [ \Delta \mathbf{x} ] )$. 

We collect $N$ samples of $M$ from a uniform distribution, so that each pattern is given an equal probability $\frac{1}{N}$ on the update. From the $N$ instances of $M(\mathbf{x})$, we select the masks which result in a correct prediction of the class. $N$ is made sufficiently large, so that the number of correct predictions $N^* > 10,000$. Next, we see that $\mathbb{E} [ \Delta \mathbf{x} ]$ is proportional to the mean of the selected updated vectors.

\begin{equation}
\begin{aligned}
     \mathbb{E} [ \Delta \mathbf{x} ] &\approx \sum_{i=1}^{N^*} Pr(M(\mathbf{x}_i)) \Delta \mathbf{x}_i \\
     &= \sum_{i=1}^{N^*} \frac{1}{N} \Delta \mathbf{x}_i \quad
     \propto  \overline{\Delta \mathbf{x}}
\end{aligned}
\end{equation}

As mentioned in equation (4), we have a normalizing step in the subsequent transform $T$. Therefore, we can calculate $\mathbb{E} [ \Delta \mathbf{x} ]$ scaled by a constant of proportionality relative to its true value, and normalize. This is given by $M(\mathbf{x}) + T\left( \ \overline{ \Delta \mathbf{x}} \ \right)$. Figure \ref{fig:generalmask} shows results probing a simple three-layered CNN model trained on MNIST\cite{lecun2010mnist}, and a ResNET model\cite{He2015DeepRL} trained on Fashion MNIST\cite{Xiao2017FashionMNISTAN}. We observe that a prototypical image for each class is obtained, irrespective of the datasets, and the type of model used.

\subsection{Progressively select $M$ from non overlapping masks}
This experiment is another method of using masks to test if a model has learnt prototypical features of an image. In this version, $M$ is drawn from a set of non overlapping masks $\mathcal{M} = \{ M_1 , M_2, ... M_k \} $. The union $ M_1 \cup M_2 \cup ... \cup M_k$ will mask out the entire image. Each mask is having the same masking ratio $\eta = \frac{1}{k}$, and thereby we can gradually visualise $T(\Delta \mathbf{x})$ for each $M_i$ selected from $\mathcal{M}$. The first row of figure \ref{fig:progressivemask} shows a set of $k=4$ masks, each masking out 25\% of the pixels. The second row of image \ref{fig:progressivemask} shows the portion of the image that is predicted(white), by iteratively masking and predicting in the preceding steps. 

Once a masking sequence is selected, we gradually predict $T_{i}(\Delta \mathbf{x})$ for each mask $M_i \in \mathcal{M}$ to reconstruct a full image. For the example shown in figure \ref{fig:progressivemask}, we take the first prediction for 25\% of the image. At this stage, 75\% of the image remains from the original image. Here we fix the predicted 25\% of pixels in the input, mask another non-overlapping region of 25\%, and predict updates for those masked pixels. After the second step of our progression, 50\% of the image is a predicted pattern, and 50\% of pixels remain from the original image. This is continued until 100\% of the pixels are from predicted updates, and 0\% of pixels are from the original image.

Formally, we keep $T_{1}(\Delta \mathbf{x})$ as the features for the masked pixels by $M_1$, and use that as part of the unmasked context to predict $T_{2}(\Delta \mathbf{x})$. This continues until all $n$ pixels are replaced with update terms $T_{i}(\Delta \mathbf{x})$. Row 3 of figure \ref{fig:progressivemask} shows this procedure for the set of masks selected. After each $T_{i}(\Delta \mathbf{x})$ is visualised, it is used to predict the next set of masked pixels $T_{i+1}(\Delta \mathbf{x})$. This probes if the model can gradually re-create a version of the original image, by conditioning on the first unmasked set of pixels. We observe that each of the predictions $T_{i}(\Delta \mathbf{x})$ contain sufficiently distinctive patterns, such that iteratively conditioning on predicted updates lead to a recognisable prototypical image of the underlying class.

\subsection{Consistency of the prototypical image}

To check if the prototypical image is consistent beyond human interpretation, we ask the question: ``Can the model recognise our visualisation of the `prototypical image', as an instance of the original class ?". Thereby, we create a new test dataset with prototypical images generated from our method of visualisation. For each image, the label is the predicted label $\hat{y}$ for the masked input. Afterwards, we test the model $f$ to see if it can classify the prototypical images correctly. Common metrics of classification accuracy such as accuracy and F1-score are reported in table \ref{table:consistency}, to demonstrate that the prototypical images are sufficiently classified by the original model itself. This confirms that our proposed prototypical image resembles the original class, both in the perspective of a human observer, and from the the perspective of the trained model $f$ itself.

\begin{table*}[h!]
    \centering
    \caption{A study on the prediction accuracy of $f$, with the test set of prototypical images. This attempts to probe if the general structure that we form from $f$, is classified correctly by $f$ itself.}
    \label{table:consistency}
    \begin{tabular}{|l|l|l|l|l|l|l|}
    \hline
        Dataset & Model$f$ & \# Classes & Masking Mechanism & accuracy & macro avg. f1-score & weighted avg. f1-score  \\ \hline
        MNIST & ResNet5 & 10 & random (4.1) & 99 & 99 & 99  \\ \hline
        MNIST & ResNet5 & 10 & progressive(4.2) & 93 & 93 & 93 \\ \hline
    \end{tabular}

\end{table*}

\subsection{Effect of the masking ratio and patch size}
\begin{figure}[htb]
\begin{minipage}[b]{1.0\linewidth}
  \centering
  \centerline{\includegraphics[width=8.5cm]{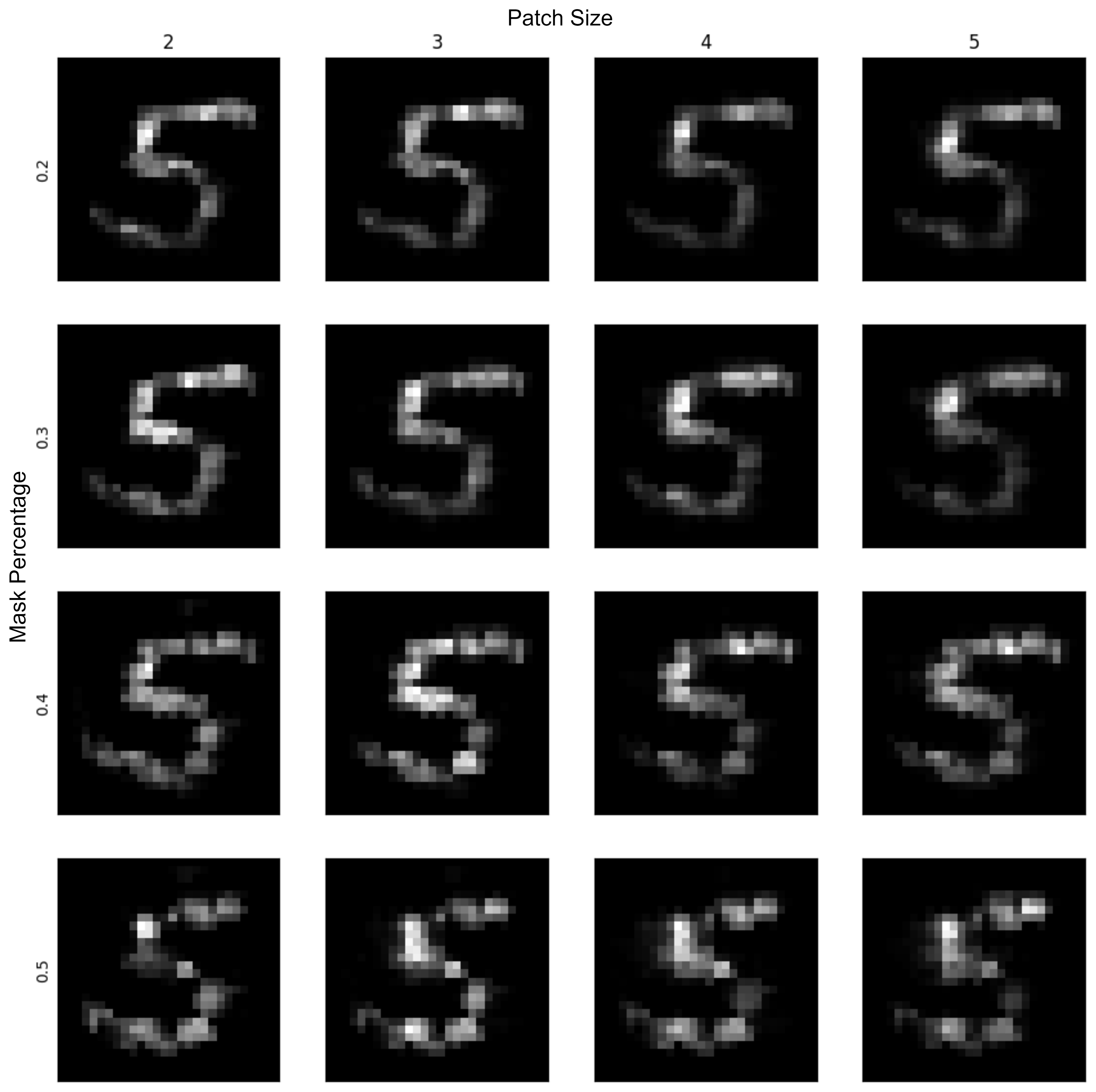}}
\end{minipage}
\caption{Visual ablation of the choice of mask ratio $\eta$ and the patch size of each mask. }
\label{fig:visualablation}
\end{figure}

For our visualisation of the update term $\Delta \mathbf{x}$, we have a thresholding function $T$ as described in equation (4). In this experiment, we visually ablate the choice of masking ratio $\eta$ and the patch size of the mask $M$. For a higher masking ratio, more pixels will be masked at the forward pass, and blanks that are filled in by the update will be less interpretable. The patch size corresponds to the number of pixels that are clumped together, forming a superpixel for each mask. For larger patch sizes, more local information is lost by masking, leading to a more granular visualisation of the update terms. In the previous experiments, patch size was set to $2 \times 2$ pixel regions, and $\eta = 0.25$.
In figure \ref{fig:visualablation}, we show the effect of other patch sizes and mask ratios on MNIST dataset. We further note that this visual ablation will have different results depending on the resolution of the images in the dataset. Therefore, the hyperparameters would have to be tuned depending on the dataset used for interpretation.

\subsection{Second order update terms}
\begin{figure}[htb]
\begin{minipage}[b]{1.0\linewidth}
  \centering
  \centerline{\includegraphics[width=8.5cm]{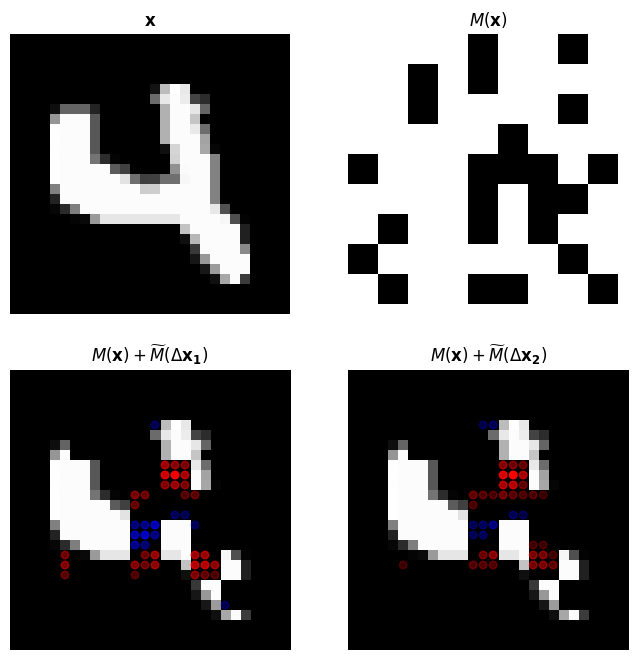}}
\end{minipage}
\caption{Comparing how the model ``fills in the blanks" from a first order gradient update term $\Delta \mathbf{x}_1$, with the predictions from the second order direction vector $\Delta \mathbf{x}_2$.}
\label{fig:gradvBFGS}
\end{figure}

As mentioned in equation (2) and section 3.2, the update term $\Delta \mathbf{x}$ need not strictly be a first order term of gradient descent. In general, any approximation of an update term in the image space, that locally searches and optimizes for the loss function $\mathcal{L}_\theta$ satisfies the conditions needed. The main aim is to interpret the changes `suggested' for the image space to ``fill in the blanks", and therefore we need to visualise the direction of the vector given by a local optimization algorithm. In this experiment, we visualise the effect of using a second order update term $\Delta \mathbf{x}_2$, which is obtained using the Broyden, Fletcher, Goldfarb, and Shanno (BFGS) algorithm for unconstrained numerical optimization. Precisely, the update direction $p_t$ is obtained by solving the equation $\mathbf{B}_t \mathbf{p}_t = - \nabla \mathcal{L}_\theta(\mathbf{x}_t)$. Here $\mathbf{B}_t$ is the approximation of the Hessian matrix at step $t$. We use the update direction $\Delta \mathbf{x} = \mathbf{p}_t$ as input to the thresholding function $T$, and compare the with the visualisations of first order gradient update $\Delta \mathbf{x}_1$. As depicted in figure \ref{fig:gradvBFGS}, there are no significant improvements by opting for a second order update for MNIST data. Therefore, we employ the first-order update term for ease of computation.

\subsection{Choice of binarization}
\begin{figure}[htb]
\begin{minipage}[b]{1.0\linewidth}
  \centering
  \centerline{\includegraphics[width=8.5cm]{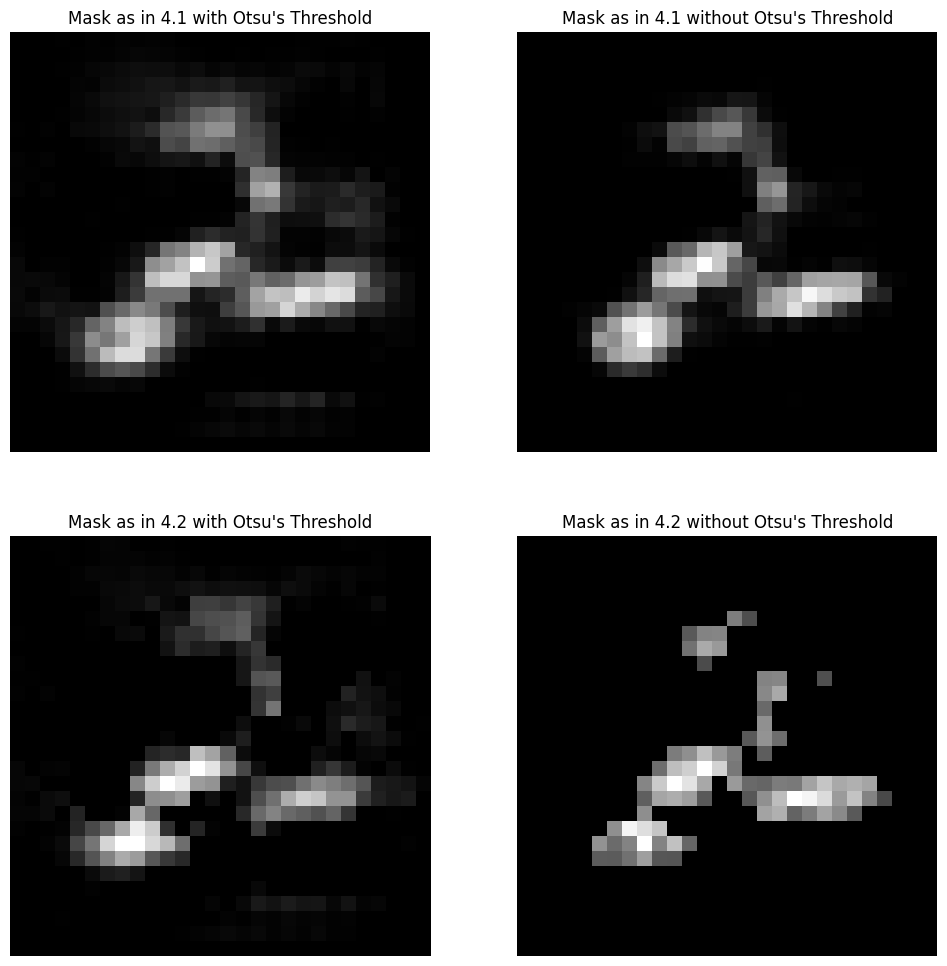}}
\end{minipage}
\caption{Comparing the effect of the binary threshold operation in $T$. The predominant features of the class are preserved in both proposed variants of masking (4.1 and 4.2). }
\label{fig:Otsu}
\end{figure}

Our choice of the binarization method, was motivated by the fact that we only wanted to visualize the most salient patterns from the update term. In figure \ref{fig:Otsu} we show the visualisations that motivated this choice. The binarized updates are more sharp with better contrast in the features, while the non-binarized version has artifacts, possibly from other classes that the model has trained on. This is because the model never makes a one-hot prediction. Therefore backpropagation from the predicted distribution $\hat{\mathbf{y}}$ contains artifacts of features from classes other than the predicted label. 



\section{Conclusions and Future work}
\label{sec:futurework}
In this work, we re-think the task of interpreting vision-based classifier models. As such, we propose to visualise how a pre-trained image classifier would `Fill in the blanks' of a masked image. We show two approaches of masking, to demonstrate ways of revealing the underlying structure of a prototypical class. Experiments of MNIST and FashionMNIST datasets reveal that our approach is able to generate visual outputs which resemble the structures that are conditioned on the masked image.

We aim to extend this idea of `Interpretability via filling the blanks', into classification more generally, and explore applications in language tasks such as sentiment analysis. Further, we aim to release a software package that automatically visualises how classifiers `fill in the blanks', and which can be easily integrated with models coded in common platforms such as pytorch and tensorflow.

\pagebreak
\bibliographystyle{IEEEbib}
\bibliography{strings,refs}

\begin{thebibliography}{10}

\bibitem{MontavonMethods}
Gr{\'{e}}goire Montavon, Wojciech Samek, and Klaus{-}Robert M{\"{u}}ller,
\newblock ``Methods for interpreting and understanding deep neural networks,''
\newblock {\em CoRR}, vol. abs/1706.07979, 2017.

\bibitem{Deconvnet_Zeiler2013}
Matthew~D. Zeiler and Rob Fergus,
\newblock ``Visualizing and understanding convolutional networks,''
\newblock {\em ArXiv}, vol. abs/1311.2901, 2013.

\bibitem{UnderstandingInverting_Mahendran2014}
Aravindh Mahendran and Andrea Vedaldi,
\newblock ``Understanding deep image representations by inverting them,''
\newblock in {\em 2015 IEEE Conference on Computer Vision and Pattern Recognition (CVPR)}, 2014, pp. 5188--5196.

\bibitem{understandvisualising_yosinski2015}
Jason Yosinski, Jeff Clune, Anh Nguyen, Thomas Fuchs, and Hod Lipson,
\newblock ``Understanding neural networks through deep visualization,''
\newblock {\em CoRR}, 2015.

\bibitem{LRP_Bach2015}
Bach S, Binder A, Montavon G, Klauschen F, Müller K-R, and Samek W,
\newblock ``On pixel-wise explanations for non-linear classifier decisions by layer-wise relevance propagation.,''
\newblock {\em PLoS ONE}, vol. 10, pp. 7, 2015.

\bibitem{LIME_Ribeiro2016}
Marco~Tulio Ribeiro, Sameer Singh, and Carlos Guestrin,
\newblock ``“why should i trust you?”: Explaining the predictions of any classifier,''
\newblock {\em Proceedings of the 22nd ACM SIGKDD International Conference on Knowledge Discovery and Data Mining}, 2016.

\bibitem{DEEPLIFT_Shrikumar2017}
Avanti Shrikumar, Peyton Greenside, and Anshul Kundaje,
\newblock ``Learning important features through propagating activation differences,''
\newblock in {\em International Conference on Machine Learning (ICML)}, 2017.

\bibitem{AxiomaticAF_Sundararajan2017}
Mukund Sundararajan, Ankur Taly, and Qiqi Yan,
\newblock ``Axiomatic attribution for deep networks,''
\newblock in {\em International Conference on Machine Learning (ICML)}, 2017.

\bibitem{CAM_Zhou2015}
Bolei Zhou, Aditya Khosla, {\`A}gata Lapedriza, Aude Oliva, and Antonio Torralba,
\newblock ``Learning deep features for discriminative localization,''
\newblock {\em 2016 IEEE Conference on Computer Vision and Pattern Recognition (CVPR)}, pp. 2921--2929, 2015.

\bibitem{GradCAM_Selvaraju2016}
Ramprasaath~R. Selvaraju, Abhishek Das, Ramakrishna Vedantam, Michael Cogswell, Devi Parikh, and Dhruv Batra,
\newblock ``Grad-cam: Visual explanations from deep networks via gradient-based localization,''
\newblock {\em International Journal of Computer Vision}, vol. 128, pp. 336 -- 359, 2016.

\bibitem{ScoreCAM_Wang2019}
Haofan Wang, Zifan Wang, Mengnan Du, Fan Yang, Zijian Zhang, Sirui Ding, Piotr~(Peter) Mardziel, and Xia Hu,
\newblock ``Score-cam: Score-weighted visual explanations for convolutional neural networks,''
\newblock {\em 2020 IEEE/CVF Conference on Computer Vision and Pattern Recognition Workshops (CVPRW)}, pp. 111--119, 2019.

\bibitem{AblationCAM_Desai2020}
Saurabh~Satish Desai and H.~G. Ramaswamy,
\newblock ``Ablation-cam: Visual explanations for deep convolutional network via gradient-free localization,''
\newblock {\em 2020 IEEE Winter Conference on Applications of Computer Vision (WACV)}, pp. 972--980, 2020.

\bibitem{EigenCAM_Fu2020}
Ruigang Fu, Qingyong Hu, Xiaohu Dong, Yulan Guo, Yinghui Gao, and Biao Li,
\newblock ``Axiom-based grad-cam: Towards accurate visualization and explanation of cnns,''
\newblock {\em ArXiv}, vol. abs/2008.02312, 2020.

\bibitem{Erhan2009VisualizingHF}
D.~Erhan, Yoshua Bengio, Aaron~C. Courville, and Pascal Vincent,
\newblock ``Visualizing higher-layer features of a deep network,''
\newblock 2009.

\bibitem{DeepInside_Simonyan2013}
Karen Simonyan, Andrea Vedaldi, and Andrew Zisserman,
\newblock ``Deep inside convolutional networks: Visualising image classification models and saliency maps,''
\newblock {\em CoRR}, vol. abs/1312.6034, 2013.

\bibitem{StrivingFS_Springenberg2014}
Jost~Tobias Springenberg, Alexey Dosovitskiy, Thomas Brox, and Martin~A. Riedmiller,
\newblock ``Striving for simplicity: The all convolutional net,''
\newblock in {\em International Conference on Learning Representations (ICLR)}, 2014.

\bibitem{VisualCNNPreImg_Mahendran2015}
Aravindh Mahendran and Andrea Vedaldi,
\newblock ``Visualizing deep convolutional neural networks using natural pre-images,''
\newblock in {\em International Journal of Computer Vision}, 2015, pp. 233--255.

\bibitem{SynthesizingTP_Nguyen2016}
Anh~M Nguyen, Alexey Dosovitskiy, Jason Yosinski, Thomas Brox, and Jeff Clune,
\newblock ``Synthesizing the preferred inputs for neurons in neural networks via deep generator networks,''
\newblock in {\em Neural Information Processing Systems}, 2016.

\bibitem{PPGN_Nguyen2017}
A.~Nguyen, J.~Clune, Y.~Bengio, A.~Dosovitskiy, and J.~Yosinski,
\newblock ``Plug and amp; play generative networks: Conditional iterative generation of images in latent space,''
\newblock in {\em 2017 IEEE Conference on Computer Vision and Pattern Recognition (CVPR)}, 2017, pp. 3510--3520.

\bibitem{lecun2010mnist}
Yann LeCun, Corinna Cortes, and CJ~Burges,
\newblock ``Mnist handwritten digit database,''
\newblock {\em ATT Labs [Online]. Available: http://yann.lecun.com/exdb/mnist}, vol. 2, 2010.

\bibitem{olah2017feature}
Chris Olah, Alexander Mordvintsev, and Ludwig Schubert,
\newblock ``Feature visualization,''
\newblock {\em Distill}, 2017,
\newblock https://distill.pub/2017/feature-visualization.

\bibitem{Xiao2017FashionMNISTAN}
Han Xiao, Kashif Rasul, and Roland Vollgraf,
\newblock ``Fashion-mnist: a novel image dataset for benchmarking machine learning algorithms,''
\newblock {\em ArXiv}, vol. abs/1708.07747, 2017.

\bibitem{He2015DeepRL}
Kaiming He, X.~Zhang, Shaoqing Ren, and Jian Sun,
\newblock ``Deep residual learning for image recognition,''
\newblock in {\em 2016 IEEE Conference on Computer Vision and Pattern Recognition (CVPR)}, 2015.

\end{thebibliography}
\end{document}